%% file: main.tex
\documentclass[letterpaper, 10 pt, conference]{ieeeconf}

\IEEEoverridecommandlockouts 
\usepackage{graphicx}
\usepackage{epsfig} 
\usepackage{amsmath} 

\usepackage{amssymb}  
\usepackage{amsthm}
\usepackage{wasysym}
\usepackage{mathrsfs}  
\usepackage{bbm}
\usepackage{color}
\usepackage[ruled,vlined,linesnumbered]{algorithm2e}
\usepackage[colorlinks=true,linkcolor=black,anchorcolor=black,citecolor=black,filecolor=black,menucolor=black,runcolor=black,urlcolor=black]{hyperref}
\usepackage[normalem]{ulem} 
\usepackage{etoolbox}
\usepackage[table,xcdraw,dvipsnames]{xcolor}
\usepackage{multirow}
\usepackage{mathtools}
\usepackage{booktabs}
\usepackage{textcase}
\usepackage{array}
\usepackage{tabularx}

\usepackage[table]{xcolor}
\definecolor{rank1}{HTML}{FF9999} 
\definecolor{rank2}{HTML}{FFCC99} 
\definecolor{rank3}{HTML}{FFFF99} 

\long\def\@makecaption#1#2{%
\ifx\@captype\@IEEEtablestring%
\begin{center}{\footnotesize #1}\\{\footnotesize #2}\end{center}%
\@IEEEtablecaptionsepspace%
\else
\@IEEEfigurecaptionsepspace%
\setbox\@tempboxa\hbox{\footnotesize #1.~~ #2}%
\ifdim \wd\@tempboxa >\hsize%
\setbox\@tempboxa\hbox{\footnotesize #1.~~ }%
\parbox[t]{\hsize}{\footnotesize \noindent\unhbox\@tempboxa#2}%
\else%
\ifcenterfigcaptions \hbox to\hsize{\footnotesize\hfil\box\@tempboxa\hfil}%
\else \hbox to\hsize{\footnotesize\box\@tempboxa\hfil}%
\fi\fi\fi}

\usepackage{float}
\usepackage{siunitx} 

\usepackage{outlines}
\let\labelindent\relax
\usepackage{enumitem}
\setenumerate[1]{label=\arabic*.}
\setenumerate[2]{label=\alph*.}

\usepackage{algorithm2e}
\RestyleAlgo{ruled}

\input{macros.tex}

\usepackage[
    style=ieee,
    doi=false,
    isbn=false,
    url=false,
    eprint=false,
    backend=biber,
    natbib=true,
    citestyle=numeric-comp,
    maxbibnames=99
    ]{biblatex}

\usepackage{cleveref}
\crefname{figure}{Fig.}{Figs.}
\Crefname{figure}{Figure}{Figures}
\Crefname{problem}{Problem}{Problems}
\Crefname{thm}{Theorem}{Theorems}
\Crefname{prop}{Proposition}{Propositions}
\Crefname{assum}{Assumption}{Assumptions}
\Crefname{lem}{Lemma}{Lemmas}
\Crefname{enumi}{Assumption}{Assumptions}

\bibliography{references}

\title{\LARGE \bf
MicroSplat: Towards Automating 3D Small-Scale Reconstruction
}

\author{
Roman Mineyev$^\dagger$,
Robert Azarcon$^\dagger$,
Suvrat Jain,
Chuizheng Kong, and
Shreyas Kousik
\thanks{$^\dagger$Equal contribution.}
\thanks{All authors are with the Georgia Institute of Technology. Corresponding author: Roman Mineyev, \texttt{rmineyev3@gatech.edu}.}
}

\usepackage[export]{adjustbox} 

\makeatletter
\def\thetable{\@arabic\c@table}
\def\fnum@table{Table~\thetable}
\long\def\@makecaption#1#2{%
\ifx\@captype\@IEEEtablestring%
\@IEEEtablecaptionsepspace%
\setbox\@tempboxa\hbox{\footnotesize #1.~~ #2}%
\ifdim \wd\@tempboxa >\hsize%
\setbox\@tempboxa\hbox{\footnotesize #1.~~ }%
\parbox[t]{\hsize}{\footnotesize \noindent\unhbox\@tempboxa#2}%
\else%
\ifcenterfigcaptions \hbox to\hsize{\footnotesize\hfil\box\@tempboxa\hfil}%
\else \hbox to\hsize{\footnotesize\box\@tempboxa\hfil}%
\fi\fi%
\else
\@IEEEfigurecaptionsepspace%
\setbox\@tempboxa\hbox{\footnotesize #1.~~ #2}%
\ifdim \wd\@tempboxa >\hsize%
\setbox\@tempboxa\hbox{\footnotesize #1.~~ }%
\parbox[t]{\hsize}{\footnotesize \noindent\unhbox\@tempboxa#2}%
\else%
\ifcenterfigcaptions \hbox to\hsize{\footnotesize\hfil\box\@tempboxa\hfil}%
\else \hbox to\hsize{\footnotesize\box\@tempboxa\hfil}%
\fi\fi\fi}
\makeatother

\begin{document}

\maketitle

\input{sections/00_abstract.tex}
\input{sections/01_introduction.tex}
\input{sections/02aa_related_work}
\input{sections/03_proposed_method.tex}

\input{sections/04_examples.tex}
\input{sections/05_conclusion.tex}

\renewcommand{\bibfont}{\normalfont\footnotesize}
{\renewcommand{\markboth}[2]{}
\printbibliography}

\end{document}

%% file: macros.tex
\DeclareSIUnit{\pixel}{px}



%% file: sections/00_abstract.tex
\begin{abstract}
Reconstructing sub-millimeter detail of small objects underpins tasks such as circuit-board defect detection, specimen digitization, and archiving.
3D Gaussian Splatting (3DGS) makes photorealistic reconstruction possible from posed images alone, but macro optics violate two conditions it depends on.
The field of view (FOV) and the depth of field (DoF) shrink to millimeters, so viewpoint error can take a sample out of frame, and only a thin focal plane is sharp in frame.
Existing automated macro-capture rigs handle the shallow DoF with per-view focus stacking, but typically rely on classical photogrammetry and offline, human-in-the-loop processing.
We present MicroSplat, which generates a 3DGS using a low-cost consumer microscope on a manipulator arm.
After a single up-front object-localization step, the system runs unattended.
It performs calibration,
plans collision-free viewpoints, captures and fuses a focal stack at each viewpoint, segments the object across all views, and optimizes a 3DGS model warm-started with the arm's own poses.
We further compare against blur-aware 3DGS methods, which infer sharp scenes from defocused images, and find that measuring sharpness with a controlled focal sweep outperforms inferring it. We provide real and synthetic datasets (including robot and camera parameters) and the fitted 3DGS reconstructions.
\end{abstract}

%% file: sections/01_introduction.tex
\section{Introduction}

Centimeter-scale objects resolved at sub-millimeter detail are the working domain of a range of inspection and digitization tasks: defect detection on printed circuit boards (PCBs) and solder joints, the digitization of insects and biological specimens,
and the archiving of coins and other small artifacts~\cite{fu2023pcbicp,lu2023solderjoint,macdonald2017coins,doan2024insectscanner}.
3D Gaussian Splatting (3DGS)~\cite{kerbl20233dgs} makes photorealistic reconstruction possible from posed RGB images alone, but macro optics break two of its working assumptions at once.
The field of view and the depth of field (DoF) both shrink to millimeters, so viewpoint error decides whether the sample is in frame at all; and within a single exposure only a thin focal plane is sharp, so recovering sharp detail requires an explicit strategy.
Two families exist: focus stacking, which recovers sharpness before reconstruction by combining multiple images per viewpoint, and blur-aware methods, which recover a sharp scene from defocused images during the reconstruction phase.
Systems that automate macro capture, such as the Orbitoscope~\cite{ylinampa2026orbitoscope}, already handle the shallow DoF with per-view focus stacking, positioning their sample by hand at a fixed orbit center. What they do not produce is a Gaussian Splat; these systems reconstruct with classical photogrammetry and rely on offline, human-in-the-loop processing.
The manual efforts that do create a Gaussian Splat require labor-intensive hand-masking of hundreds of images to suppress floater artifacts (spurious semi-transparent Gaussians) induced by blurred backgrounds~\cite{bittel2025macro}.

\begin{figure}[t]
  \centering
  \includegraphics[width=\columnwidth]{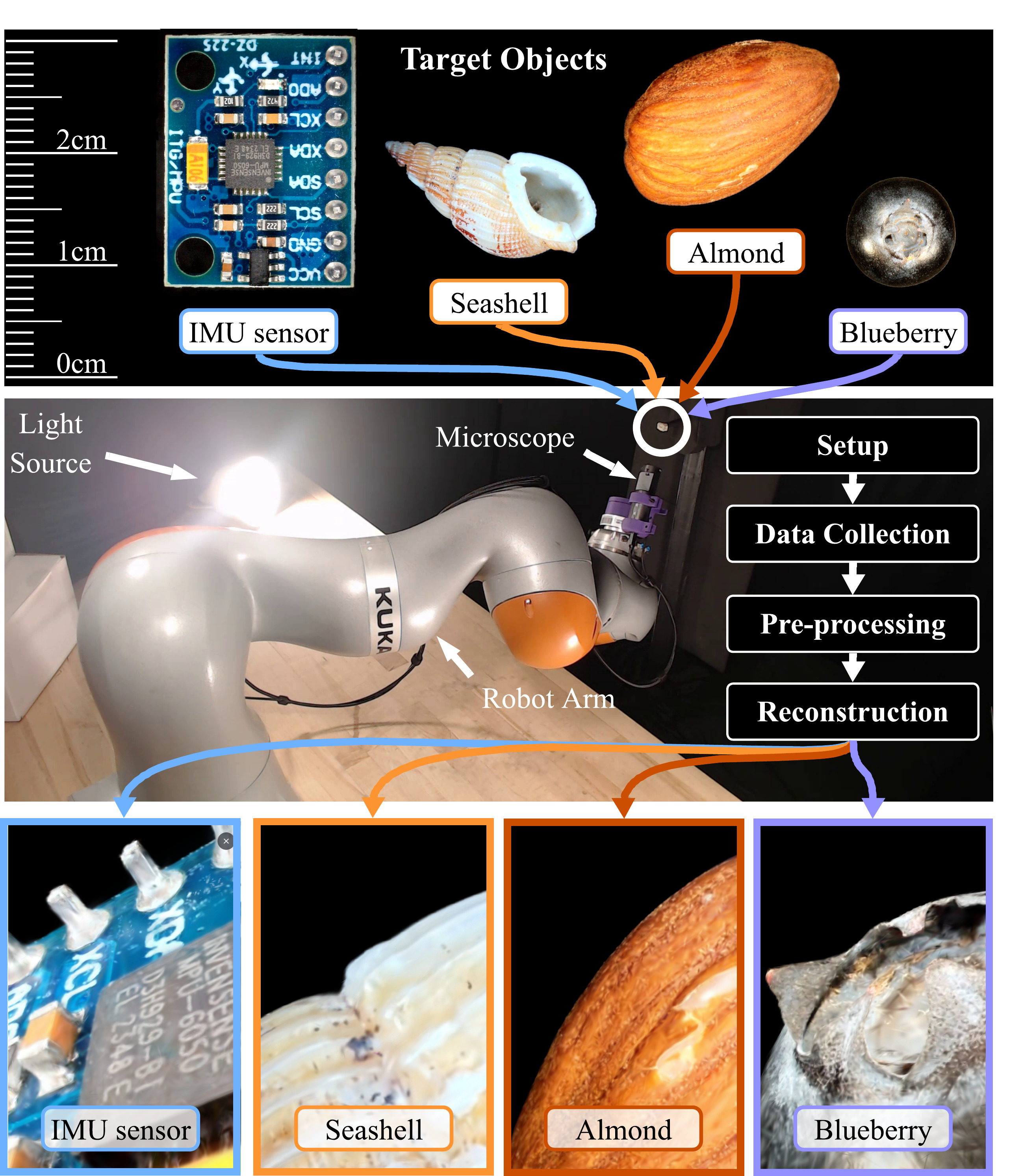}
  \vspace*{-0.5cm}
  \caption{Overview of our proposed MicroSplat pipeline for automatically generating photorealistic reconstructions of centimeter-scale objects at sub-millimeter detail.}
  \label{fig:hero}
  \vspace*{-0.3cm}
\end{figure}


MicroSplat\footnote{Project page (includes interactive viewer): \url{https://microsplat.github.io/}. Code: \url{https://anonymous.4open.science/r/Microscopic-3D-Reconstruction-FA3D}} pairs a robotic manipulator arm (KUKA iiwa14 R820) with a low-cost consumer microscope as its end-effector (\cref{fig:hero}).
We leverage the arm's \SI{\pm 0.1}{\milli\meter} repeatability to enable microscope calibration and precise microscope placement for focus stacking.
Our use of a low-cost microscope introduced noise in both our images and poses. We developed several engineering solutions such as a human-in-the-loop object-centering pipeline, SfM initialization with robot pose and camera intrinsics, and Gaussian-fit-based window selection for focus stacking to make our pipeline robust to this noise.

To summarize, we make three main contributions:

\begin{enumerate}
  \item \textbf{An autonomous capture-to-reconstruction pipeline}, in which, after calibration and object-centering, the system captures data and generates a 3DGS model of a small object.
  We also extensively evaluate the impact of key engineering choices in our pipeline (object centering and robot pose SfM initialization) through ablations and comparison to baselines.

  \item \textbf{A comparison against blur-aware 3DGS methods}, in which we benchmark our focus-stacking approach on synthetic data using ground-truth in-focus renderings, and show that when a controlled robotic focal sweep captures sharp detail from every viewpoint, measuring sharpness directly outperforms inferring it from defocused images during reconstruction.
    
  \item \textbf{A reconstruction dataset}
    of eight real objects and eight synthetic objects, spanning a wide range of attributes. The dataset contains information on scans, 3DGS reconstructions, initial robot arm poses and camera intrinsics. For each synthetic object we additionally release the Blender scene and object mesh, along with both the shallow-DoF focal stacks and the corresponding fully in-focus renderings, which serve as ground truth.\footnote{\url{https://huggingface.co/datasets/splatmicro/microsplat}, released under CC BY 4.0.} 
\end{enumerate}

The rest of the paper is organized as follows:
\Cref{sec:related} covers related work, \Cref{sec:method} details our method's hardware, calibration, planning,
and reconstruction stack, and \Cref{sec:experiments} presents experimental results.

\begin{figure}[b]
\centering
\includegraphics[width=\columnwidth]{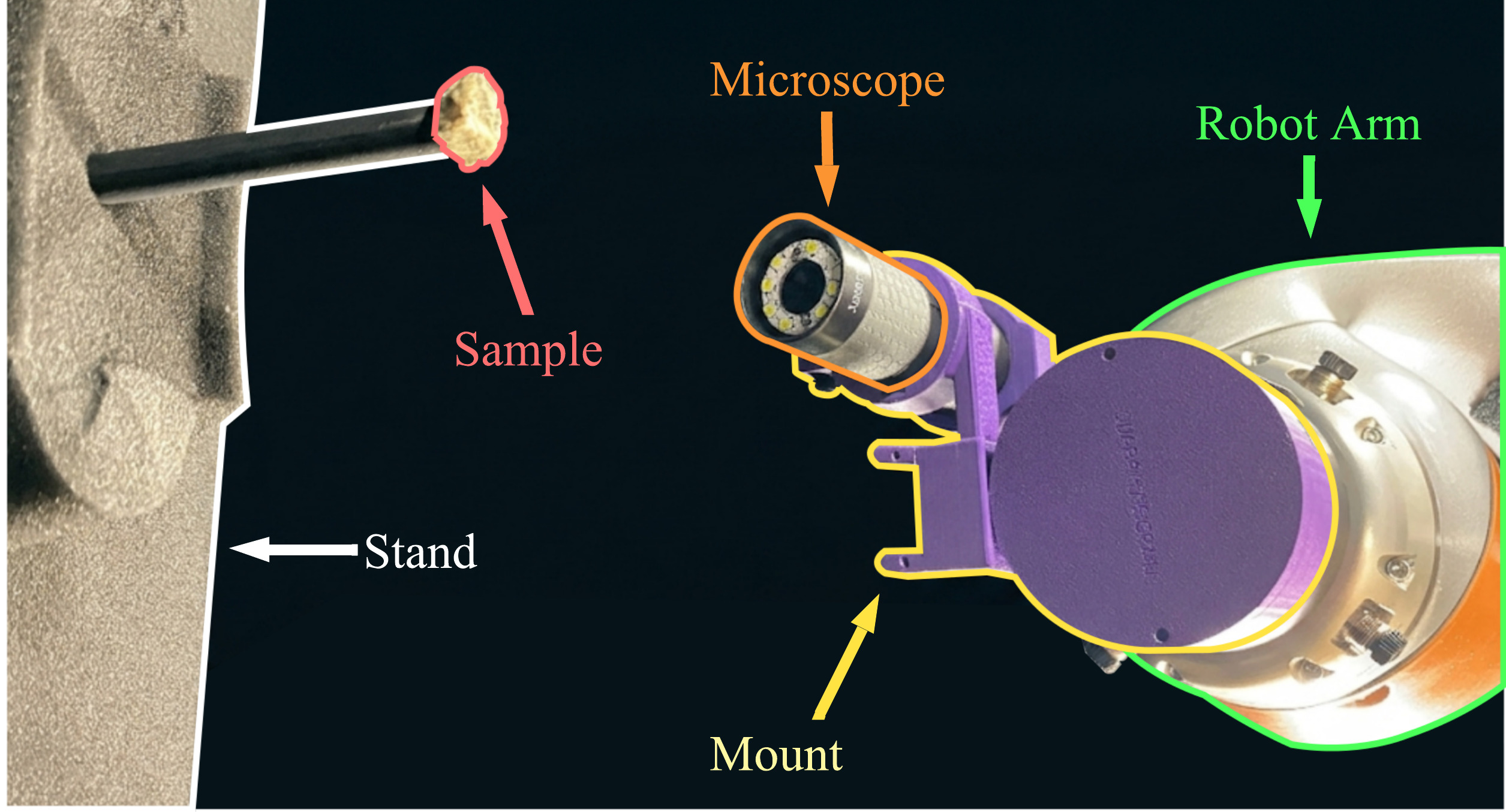}
\vspace*{-0.5cm}
\caption{Hardware setup. (Left) Sample mounted onto the stand and (right) KUKA iiwa14 R820 with custom microscope end-effector.}
\label{fig:microsplat-hw}
\vspace*{-0.3cm}
\end{figure}

\begin{figure*}[t]
  \centering
  \includegraphics[width=\textwidth]{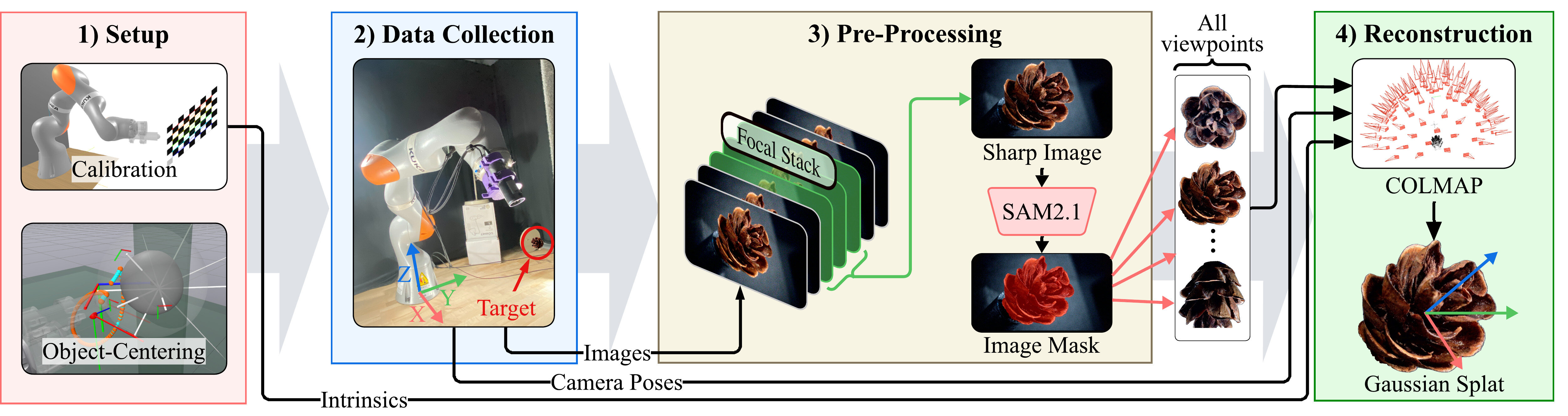}
  \vspace*{-0.5cm}
  \caption{
  Overview of our proposed method.
  Note the checkerboard in the calibration step is not drawn to scale.}
  \label{fig:architecture}
  \vspace*{-0.4cm}
\end{figure*}

%% file: sections/02aa_related_work.tex
\section{Related Work}
\label{sec:related}

We now review existing systems for automated 3D capture, highlighting the gap in small-scale capture that our system targets.

\subsection{Automated Capture Rigs}
3D reconstruction is an essential tool used across a wide range of domains such as film, video games, virtual and augmented reality, medical imaging, and robotics.
Neural rendering has emerged as a powerful approach to this problem, with Neural Radiance Fields (NeRF)~\cite{mildenhall2020nerf} and 3DGS~\cite{kerbl20233dgs} being commonly adopted over classical photogrammetry for their ability to produce photorealistic novel-view synthesis from posed images alone.

With so many applications relying on 3D reconstruction, much work has focused on automating the data collection process.
For example, the Google Scanned Objects dataset demonstrated a custom enclosure with structured light and machine vision cameras capable of generating high-fidelity 3D models for over 1,000 household items for the downstream task of robot learning~\cite{downs2022google}.

More recently, automated pipelines have moved toward reconstructing objects directly with 3DGS from RGB images alone, without the depth or structured-light hardware these earlier systems require~\cite{qiu2025omniscan, strong2024nextbestsense, jiang2024fisherrf, pan2022activenerf}.
For example, Omni-Scan~\cite{qiu2025omniscan} used a bi-manual robot to rotate a grasped object in front of a fixed RGB camera, merging two scans into a 3DGS model.
To isolate the object from the gripper and background, it prompts SAM~2~\cite{ravi2025sam2} with foreground cues and scores the resulting candidate masks.
We similarly use SAM 2 for object mask generation in MicroSplat.

\subsection{Small-Scale Reconstruction}
Beyond standard objects, there is great interest in applying 3D reconstruction to resolve sub-millimeter detail as well.
In electronics and semiconductor manufacturing, high-fidelity reconstructions of PCBs, microcontrollers, and solder joints enable automated defect detection and quality control~\cite{fu2023pcbicp, lu2023solderjoint}.
Similar techniques are used to digitize insects and biological specimens in entomology, to archive coins, engravings, and other small artifacts in cultural heritage, and to characterize surface morphology in materials science~\cite{macdonald2017coins, gallo2014multifocus, doan2024insectscanner, spp_gaussian_splatting_2024}.

\subsection{Small-Scale Capture and Sharpness Recovery}
Hobbyists such as Danny Bittel have shown that Gaussian Splatting of small objects is possible with manual macro photography~\cite{bittel2025macro}, using focus stacking~\cite{wang2011focusstacking} to overcome shallow DOF: a series of images are captured from a single viewpoint while the camera steps along the optical axis, and the resulting slices are fused into one all-in-focus composite.

Manually producing such a capture, however, faces two bottlenecks.
First, at macro working distances, both FOV and DOF are small, so precise viewpoint control is essential.
Second, masks are necessary since blurred backgrounds lack geometric cues, causing background-colored floater artifacts that forced Bittel to hand-mask hundreds of images.

Several works do automate this capture process for small samples via focus stacking~\cite{gallo2014multifocus, latapie2024pearls, ylinampa2026orbitoscope}.
Closest to this work, Orbitoscope~\cite{ylinampa2026orbitoscope} uses a custom six-axis gantry with a dedicated stacking axis for per-view focal stacks.
Its viewpoints are fixed and the sample is manually positioned at the orbit center with intersecting laser pointer.
We take the complementary approach of leaving the sample untouched and adjusting the viewpoints (i.e., the robot's motion) to the estimated sample center.

All of these autonomous systems for small-scale objects, however, reconstruct with classical photogrammetry rather than Gaussian Splatting.
To our knowledge, MicroSplat is the first fully autonomous pipeline for small-scale objects that goes from robotic data collection to a 3DGS model without human intervention after setup.


An alternative to capture-time focus stacking is to recover sharpness during reconstruction: Deblurring 3DGS~\cite{lee2024deblurring3dgs} and its successors~\cite{peng2024bags, lee2025cocogaussian} jointly estimate a blur model and a sharp scene from defocused images. However, with a robot arm we can capture in-focus detail directly from every viewpoint rather than inferring it. We compare these methods in Section~\ref{sec:comparisons}.

%% file: sections/03_proposed_method.tex
\section{Proposed Method}
\label{sec:method}

\begin{figure}[t]
  \centering
  \includegraphics[width=\columnwidth]{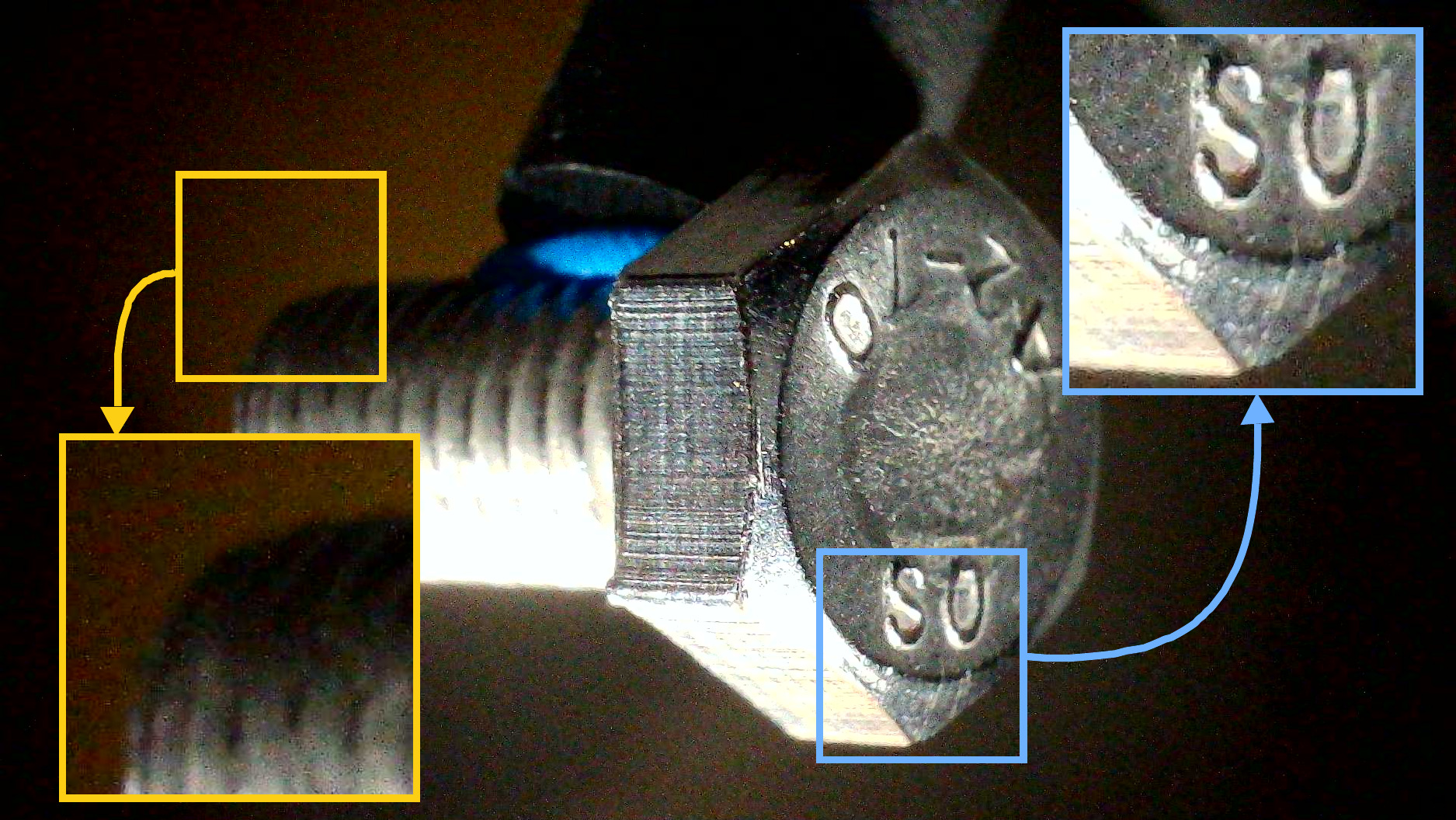}
  \vspace*{-0.5cm}
  \caption{A bolt imaged during our data collection. The letters are in focus, but the threads are blurry due to the microscope's shallow depth of field.}
  \label{fig:defocus}
  \vspace*{-0.4cm}
\end{figure}

We address the problem of producing accurate, high-fidelity renderings of small-scale objects from images captured with a digital microscope.
At this scale, the shallow DoF prevents any single image from being fully in focus (\cref{fig:defocus}), and the capture precision required is impractical to achieve manually.
We present a robotic pipeline that autonomously performs image acquisition and 3D Gaussian Splatting reconstruction.
To proceed, we first describe our hardware, then our reconstruction pipeline, which has four steps (\cref{fig:architecture}): setup, data collection, pre-processing, and reconstruction.


\subsection{Hardware}

\subsubsection{Robot}
We use a seven-degree-of-freedom KUKA iiwa14 R820 arm fitted with a custom end-effector that mounts an optical microscope (\cref{fig:microsplat-hw}).
The iiwa14 provides \SI{\pm 0.1}{\milli\meter} pose repeatability, which is essential for keeping small samples within the microscope's narrow field of view and shallow DoF across all viewpoints.

\subsubsection{Sample}
We place the sample on a fixed stand pointing towards the $-$X axis of the robot's base frame (see \cref{fig:architecture} for the robot's coordinate frame) to maximize the spherical coverage of viewpoints, as checked with standard reachability analysis.

\subsubsection{Microscope}
We use a low-cost hobbyist microscope (35 USD on Amazon at time of writing) that uses a GalaxyCore GC2063 1/2.9\textquotedblright{} RGB Bayer CMOS sensor with a resolution of $1920\times1080$ (2 MP) and a pixel pitch of \SI{2.8}{\micro\meter}.
The microscope has a zoom knob, which we kept fixed (we leave changing the zoom level to future work).
We used a calibration target of known dimensions (\SI{2}{\milli\meter} checkerboard squares with $9 \times 6$ interior corners) to calculate the object-space sampling of approximately \SI{16.44}{\micro\meter / \pixel} and FOV of  $31.57 \times 17.76$ \SI{}{\milli\meter}.
The microscope's camera image parameters, such as exposure and white balance, cannot be manually controlled.
Because 3DGS requires photometric consistency across all views, we controlled the illumination of our workspace instead, using four \SI{800}{\lumen}, \SI{5000}{\kelvin} lamps, evenly spread out behind the robot arm.



\subsection{Setup}
This first step is the pipeline's only human-in-the-loop stage, comprising a calibration phase performed once and an object-centering phase repeated for every new sample.

\subsubsection{Calibration}
We used a $9\times6$ \SI{2}{\milli\meter}-square inkjet-printed checkerboard pattern as the calibration target.
Our use of a robot arm solves the difficult of microscope calibration: the high precision needed to keep a calibration target both in the FOV and fully in-focus.
We generated viewpoints and planned the robot's motion as detailed below in \Cref{subsec: data collection}, but without collecting focal stacks.
Instead, at each viewpoint, we manually jogged the robot arm along any of the three camera-frame axes in \SI{1}{\milli\meter} increments till the calibration target was in-focus as possible.
The robot flange pose at each viewpoint was stored for hand-eye calibration.
Intrinsics calibration yielded $f_x = 6488.95$, $f_y = 6196.35$, $c_x = 865.83$, $c_y = 565.52$ (all in \si{\pixel}) with a re-projection error of \SI{0.2161}{\pixel}, and Daniilidis hand-eye calibration~\cite{daniilidis1999hand} yielded a rotation residual of \SI{0.00901}{\radian} and a translation residual of \SI{1.0707}{\milli\meter}.

\subsubsection{Human-in-the-Loop Object-Centering}
We assume the sample stays static during data collection. However, without an accurate initial location of the sample, it may be off-center or even entirely outside the image frame during data collection.

We address this with a manual object centering step.
We start with a rough guess of the sample location (either known from a previous sample, or by moving the robot manually).
We then compute five poses to use for centering the object.
The first pose aligns the camera's optical axis with the robot base X-axis. 
The remaining four poses tilt the camera up, left, and right \SI{60}{\degree}, and downward \SI{45}{\degree} (to avoid collision with table), relative to the estimated object center.

The robot plans motion to each pose with RRT* (implemented in Drake \cite{drake}).
Then, the user jogs the camera in \SI{1}{\milli\meter} increments until the sample is centered on a crosshair (shown in the user's camera feed), denoting $(c_x, c_y)$ coordinates calculated from calibration.
The 3D ray in camera coordinates passing through the pixel at $(u, v)$ is
\begin{equation}
    \mathbf{d} = \begin{bmatrix} (u - c_x) / f_x \\ (v - c_y) / f_y \\ 1 \end{bmatrix}.
\end{equation}
By manually aligning the sample to $(c_x, c_y)$, this ray is simplified to $[0, 0, 1]^T$.
The ray in world coordinates is then obtained as
\begin{equation}
    \mathbf{d}_w = \mathbf{R}_w \, \mathbf{d}, \qquad \mathbf{o}_w = \mathbf{t}_w
\end{equation}
where $\mathbf{R}_w$ and $\mathbf{t}_w$ are the rotation and translation of the camera in world coordinates.
The 3D center estimate is then found by minimizing the sum of squared distances to all five rays via least squares:
\begin{equation}
    \mathbf{p}^* = \Big( \sum_i (\mathbf{I} - \hat{\mathbf{d}}_i \hat{\mathbf{d}}_i^T) \Big)^{-1} \sum_i (\mathbf{I} - \hat{\mathbf{d}}_i \hat{\mathbf{d}}_i^T) \mathbf{o}_i,
\end{equation}
where $\hat{\mathbf{d}}_i$ is the unit direction of ray $i$ and $\mathbf{o}_i$ is its origin.

\subsection{Data Collection}\label{subsec: data collection}

We first perform offline planning (viewpoint and robot joint position generation), then do online data collection via motion planning.

\subsubsection{Offline planning}
We create candidate camera viewpoints distributed on a hemisphere centered on the object, with its axis is along the $-$X axis of the robot base frame.
We restrict coverage to a hemisphere rather than a full sphere because the arm's joint limits prevent the camera from reaching the far side of the sample, and because the stand itself physically blocks a \SI{360}{\degree} view. As a result, the backside of the sample is never observed.
Viewpoints are selected from a Fibonacci lattice~\cite{gonzalez2010fibonacci} for near-uniform angular coverage. 

Camera orientations are assigned by a look-at construction: the camera's optical axis (Z-axis) follows the outward surface normal of the hemisphere, and the camera's X-axis is the projection of the world +Z onto the hemisphere surface.
The viewpoints are then re-ordered via a nearest-neighbor greedy tour on 3D distance to minimize distance traveled for the robot arm between the viewpoints.

The orientation produced by the hemisphere sampling aims the camera's optical
axis ($\mathbf{e}_z$) at the sample.
However, the sample should appear at the
image center $(W/2,\,H/2)$, where $W$ and $H$ are the image width and height in pixels, rather than at the principal point $(c_x, c_y)$, to ensure that majority of an sample is seen in every captured image frame.
We therefore apply a small in-plane correction to each viewpoint orientation. Let $\mathbf{K}$ be the camera intrinsics.
The viewing direction in the camera
frame corresponding to the image-center pixel is
\begin{equation}
  \mathbf{d} = \frac{\mathbf{K}^{-1}\,\tilde{\mathbf{p}}_c}
                    {\lVert \mathbf{K}^{-1}\,\tilde{\mathbf{p}}_c \rVert},
  \qquad
  \tilde{\mathbf{p}}_c = \begin{bmatrix} W/2 \\ H/2 \\ 1 \end{bmatrix}.
\end{equation}
We then construct the rotation $\mathbf{R}_\text{off}$ that aligns this direction with the optical axis $\mathbf{e}_z = [0,0,1]^\top$.
Its angle and axis are
\begin{equation}
  \theta = \arccos\!\big(\mathbf{d}\cdot\mathbf{e}_z\big),
  \qquad
  \hat{\mathbf{a}} = \frac{\mathbf{d}\times\mathbf{e}_z}
                          {\lVert \mathbf{d}\times\mathbf{e}_z \rVert}.
\end{equation}
Then Rodrigues' formula yields
\begin{align}
    \mathbf{R}_\text{off} = \mathbf{I}
    + \sin\theta\,[\hat{\mathbf{a}}]_\times
    + (1-\cos\theta)\,[\hat{\mathbf{a}}]_\times^{2},  
\end{align}
where $[\hat{\mathbf{a}}]_\times$ is the skew-symmetric matrix of
$\hat{\mathbf{a}}$.
Finally, each viewpoint orientation
$\mathbf{R}_\text{wp}$ is corrected in the camera frame:
\begin{equation}
  \mathbf{R}_\text{new} = \mathbf{R}_\text{wp}\,\mathbf{R}_\text{off}.
\end{equation}
The camera positions on the hemisphere are left unchanged; only the orientation is tilted so that the sample  is centered in the image frame.

Using $\mathbf{R}_\text{off}$ and the eye-in-hand transform from the calibration step allows us to find the desired end-effector pose for each viewpoint.
These end-effector poses are converted to robot joint values via the stereographic Shoulder-Elbow-Wrist (SEW) inverse kinematics (IK) method~\cite{elias2025redundancy}, which allows us to enumerate all IK solutions of the 7-DOF KUKA robot arm analytically.
We set the SEW angle to \SI{135}{\degree}, which yields eight kinematically feasible robot joint solutions per viewpoint.
We then filter each joint solution to avoid collisions with the table and  sample, and joint limits, and remove any viewpoints that have no feasible IK solution.
Small differences in the localized sample location leave us with either 134 or 135 reachable viewpoints.

\subsubsection{Online data collection}
The robot traverses between each viewpoint via the Rapidly-exploring Random Tree* (RRT*) algorithm~\cite{karaman2011sampling} through a path that avoids any collisions, implemented in Drake~\cite{drake}.
At each viewpoint, the robot takes 30 images for the focal stack, jogging the robot a total of \SI{40}{\milli\meter} closer to the sample along the camera's optical axis.
The robot pauses \SI{0.5}{\second} before taking an image to avoid motion blur.
Additionally, the robot stops for \SI{2}{\second} when moving between viewpoints to ensure the exposure and white balance stabilize before taking pictures.

\subsection{Pre-Processing}

For accurate 3D reconstruction, we opted to keep the 3DGS method standard and instead clean up the image data.
To do so, we create an automated focus stacking pipeline to select which images to stack together, and a strategy for autonomously masking the focus-stacked image.


\subsubsection{Focus stacking}
Not all images taken in the focal stack have high sharpness content, and attempting to stack the entire focal stack could result in distorted images. 
Additionally, selecting too large a range of images for the focal stack may result in artifacts due to focus breathing. 
Therefore, for every focal stack, it is necessary to select only a subset of the sharpest images.

Due to the poor image quality of the camera and uncontrollable camera image parameters, we cannot use the typical thresholding techniques used in works such as Orbitoscope~\cite{ylinampa2026orbitoscope}.
We found that the Tenengrad sharpness score of a sequence of 30 images tends to be roughly Gaussian, and an additional pre-processing step allowed for robust window selection within our focal stack.

For a given viewpoint and its sequence of 30 images, we first apply 2D spatial median filtering ($5\times5$ kernel) to each image, since salt-and-pepper noise from the low-cost sensor creates spurious high-gradient pixels that spike the Tenengrad score independent of actual focus.
A scalar Tenengrad sharpness score $T_i$ is then computed for each image $i \in \{1, \dots, 30\}$ as the sum of squared horizontal and vertical Sobel gradient responses $G_x(x,y)$ and $G_y(x,y)$ over all pixel coordinates $(x,y)$ in the image domain $\Omega$:

\begin{equation}
    T_i = \sum_{(x,y) \in \Omega} \left( G_x(x,y)^2 + G_y(x,y)^2 \right)
\end{equation}

While spatial filtering addresses in-frame noise, the resulting sequence of sharpness scores $\{T_1, T_2, \dots, T_{30}\}$ across the focal stack can still exhibit minor fluctuations and outliers. To smooth this score distribution across the image sequence, we apply a subsequent 1D median filter (kernel size 3) directly to the sequence of scalar scores.

To model the sharpness profile across the focal stack, we fit a four-parameter Gaussian function $f(x)$ over the image indices $x \in [1, 30]$:
\begin{equation}
    f(x; a, \mu, \sigma, c) = a \exp\left( -\frac{1}{2}\left(\frac{x - \mu}{\sigma}\right)^2 \right) + c
\end{equation}
where $a$ represents the peak amplitude, $\mu$ the focal peak position, $\sigma$ the depth-of-field spread, and $c$ a baseline offset. The model parameters are estimated from the filtered sharpness scores using bounded nonlinear least-squares optimized via the Trust Region Reflective algorithm~\cite{branch1999trust} (\texttt{scipy.optimize.curve\_fit}~\cite{2020SciPy-NMeth}).

Images within one standard deviation ($\mu \pm \sigma$) of the estimated focal peak are selected for fusion. Taking the middle image of this subset as the reference frame, we apply complex wavelet-based Extended Depth-of-Field (EDOF)~\cite{forster2004complex} to generate the final focus-stacked image.

\subsubsection{Masking}

Isolating the sample from the out-of-focus background is crucial to mitigate floater artifacts during 3DGS optimization. To achieve this without manual annotation, we first construct a mask using classical edge cues at a single, well-conditioned viewpoint, then  propagate it to every other viewpoint with SAM~2~\cite{ravi2025sam2}, whose video mode conditions on this seed mask and generalizes far better than edge detection once the viewpoint changes.

\begin{figure}[tb]
\centering
\includegraphics[width=\columnwidth]{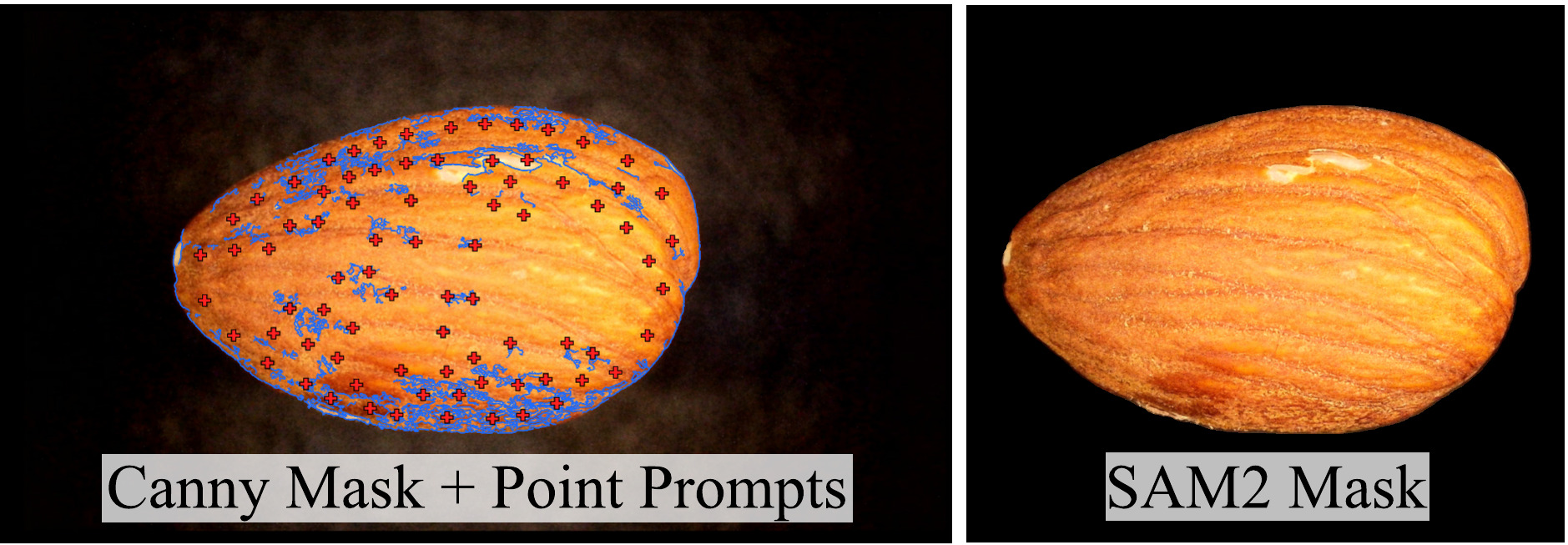}
\vspace*{-0.5cm}
\caption{Example point-prompt generation for SAM~2 segmentation. (Left) Canny
edge detection mask in blue, with sampled point prompts in red are used
to query SAM~2. (Right) Resulting segmentation mask produced by
SAM~2.}
\label{fig:masking-example}
\vspace*{-0.3cm}
\end{figure}

At the first focus-stacked image, the sample is well-centered and the mounting pole is occluded, making classical edges reliable: we extract edges via a $5\times5$ Gaussian blur followed by Canny edge detection~\cite{canny1986} and densely sample $n=80$ edge points as foreground prompts for SAM~2 in image-prediction mode. The resulting mask is slightly dilated and flood-filled, then used to seed SAM~2 in video mode (\cref{fig:masking-example}). This handoff is necessary because, at other viewpoints, the pole becomes visible and lighting shifts, causing edge-based segmentations to fail.

\begin{table*}[t]
\centering
\resizebox{\textwidth}{!}{%
\begin{tabular}{l|rrr|rrr|rrr|rrr|rrr|rrr|rrr|rrr}
& \multicolumn{3}{c|}{Almond} & \multicolumn{3}{c|}{Blueberry} & \multicolumn{3}{c|}{Bolt} & \multicolumn{3}{c|}{Circuit} & \multicolumn{3}{c|}{Flower} & \multicolumn{3}{c|}{Pinecone} & \multicolumn{3}{c|}{Rock} & \multicolumn{3}{c}{Seashell} \\
Method & PSNR$\uparrow$ & SSIM$\uparrow$ & LPIPS$\downarrow$ & PSNR$\uparrow$ & SSIM$\uparrow$ & LPIPS$\downarrow$ & PSNR$\uparrow$ & SSIM$\uparrow$ & LPIPS$\downarrow$ & PSNR$\uparrow$ & SSIM$\uparrow$ & LPIPS$\downarrow$ & PSNR$\uparrow$ & SSIM$\uparrow$ & LPIPS$\downarrow$ & PSNR$\uparrow$ & SSIM$\uparrow$ & LPIPS$\downarrow$ & PSNR$\uparrow$ & SSIM$\uparrow$ & LPIPS$\downarrow$ & PSNR$\uparrow$ & SSIM$\uparrow$ & LPIPS$\downarrow$ \\
\hline
Vanilla 3DGS & \cellcolor{rank3} 27.58 & \cellcolor{rank3} 0.65 & 0.20 & \cellcolor{rank3} 25.24 & \cellcolor{rank3} 0.75 & \cellcolor{rank3} 0.29 & \cellcolor{rank3} 22.70 & \cellcolor{rank2} 0.59 & \cellcolor{rank3} 0.17 & \cellcolor{rank3} 18.18 & \cellcolor{rank3} 0.48 & 0.16 & \cellcolor{rank3} 23.68 & \cellcolor{rank3} 0.74 & \cellcolor{rank3} 0.18 & \cellcolor{rank3} 23.47 & \cellcolor{rank3} 0.59 & \cellcolor{rank3} 0.25 & \cellcolor{rank3} 30.63 & \cellcolor{rank3} 0.77 & \cellcolor{rank3} 0.22 & \cellcolor{rank3} 28.03 & \cellcolor{rank3} 0.86 & 0.16 \\
CoCoGaussian & 27.31 & 0.64 & \cellcolor{rank3} 0.19 & 24.57 & 0.71 & 0.30 & 18.72 & 0.39 & 0.20 & 17.97 & \cellcolor{rank3} 0.48 & \cellcolor{rank3} 0.15 & 20.02 & 0.62 & 0.19 & 18.60 & 0.51 & 0.30 & 29.51 & 0.75 & \cellcolor{rank3} 0.22 & 27.01 & 0.83 & \cellcolor{rank3} 0.15 \\
Deblurring 3DGS & 26.32 & 0.62 & 0.21 & 23.30 & 0.66 & \cellcolor{rank3} 0.29 & 18.83 & \cellcolor{rank3} 0.41 & 0.21 & 17.53 & 0.46 & 0.17 & 22.78 & \cellcolor{rank3} 0.74 & \cellcolor{rank3} 0.18 & 19.25 & 0.51 & 0.29 & 29.93 & \cellcolor{rank3} 0.77 & \cellcolor{rank3} 0.22 & 24.64 & 0.79 & 0.17 \\
BAGS & \cellcolor{rank2} 29.48 & \cellcolor{rank2} 0.70 & \cellcolor{rank2} 0.17 & \cellcolor{rank2} 27.21 & \cellcolor{rank2} 0.79 & \cellcolor{rank2} 0.26 & \cellcolor{rank2} 22.78 & \cellcolor{rank2} 0.59 & \cellcolor{rank2} 0.16 & \cellcolor{rank2} 19.77 & \cellcolor{rank2} 0.56 & \cellcolor{rank2} 0.14 & \cellcolor{rank2} 26.59 & \cellcolor{rank2} 0.79 & \cellcolor{rank2} 0.16 & \cellcolor{rank2} 26.29 & \cellcolor{rank2} 0.68 & \cellcolor{rank2} 0.20 & \cellcolor{rank2} 32.70 & \cellcolor{rank2} 0.80 & \cellcolor{rank2} 0.19 & \cellcolor{rank2} 30.64 & \cellcolor{rank2} 0.89 & \cellcolor{rank2} 0.14 \\
\textbf{Ours} & \cellcolor{rank1} \textbf{33.49} & \cellcolor{rank1} \textbf{0.96} & \cellcolor{rank1} \textbf{0.06} & \cellcolor{rank1} \textbf{37.45} & \cellcolor{rank1} \textbf{0.98} & \cellcolor{rank1} \textbf{0.02} & \cellcolor{rank1} \textbf{36.06} & \cellcolor{rank1} \textbf{0.97} & \cellcolor{rank1} \textbf{0.04} & \cellcolor{rank1} \textbf{39.53} & \cellcolor{rank1} \textbf{0.99} & \cellcolor{rank1} \textbf{0.01} & \cellcolor{rank1} \textbf{32.17} & \cellcolor{rank1} \textbf{0.95} & \cellcolor{rank1} \textbf{0.06} & \cellcolor{rank1} \textbf{41.68} & \cellcolor{rank1} \textbf{0.99} & \cellcolor{rank1} \textbf{0.01} & \cellcolor{rank1} \textbf{45.39} & \cellcolor{rank1} \textbf{0.99} & \cellcolor{rank1} \textbf{0.01} & \cellcolor{rank1} \textbf{45.13} & \cellcolor{rank1} \textbf{1.00} & \cellcolor{rank1} \textbf{0.01} \\
\end{tabular}%
}
\vskip\abovecaptionskip\relax
\caption{Quantitative comparisons across all samples in the synthetic dataset. Top three per metric are colored red, orange, and yellow respectively.}
\label{tab:method-comparison}
\end{table*}

\subsection{Reconstruction}
Our reconstruction consists of COLMAP and 3DGS processes, both adjusted for better accuracy.

\subsubsection{Bundle Adjustment}

We use COLMAP~\cite{schoenberger2016sfm, schoenberger2016mvs} to refine the camera intrinsics and extrinsics, and to generate a sparse point cloud for initializing the 3DGS optimization. 
First, SIFT~\cite{lowe2004sift} features are extracted and exhaustively matched across all views.
COLMAP's incremental mapper is then seeded with our initial camera intrinsics and the robot's reported end-effector poses.
We let COLMAP refine these further, accounting for errors in intrinsics, as well as small, unmodeled pose offsets from the custom microscope mount.

\subsubsection{3D Gaussian Splatting}

We use the \texttt{gsplat} library's~\cite{ye2025gsplat} implementation of 3DGS~\cite{kerbl20233dgs}. 
To suppress floater artifacts while preserving fine structural details, we adjust the default densification hyperparameters, increasing the opacity pruning threshold to $\tau_\alpha = 0.10$ and the 2D gradient growth threshold to $\tau_\nabla = 4 \times 10^{-4}$.

Additionally, we introduce a background alpha regularization term that
penalizes rendered opacity in pixels outside the generated sample masks:
\begin{equation}
  \mathcal{L}_{\mathrm{bg}}
  = \frac{1}{|\mathcal{B}|} \sum_{(u,v)\,\in\,\mathcal{B}} \hat{\alpha}_{uv},
  \label{eq:bg_loss}
\end{equation}
where $\mathcal{B}$ represents the set of background pixels defined by our segmentation masks, and $\hat{\alpha}_{uv}$ is the rendered alpha value at pixel location $(u,v)$. This background drives the opacity of background Gaussians toward zero, ensuring most floaters are removed during the pruning phase. We set the background loss scaling factor $\lambda_{bg} = 0.2$.

Combined with the standard 3DGS photometric objective, the complete loss minimized during training is:
\begin{equation}
  \mathcal{L}
  = (1 - \lambda_{\mathrm{ssim}})\,\mathcal{L}_{1}
  + \lambda_{\mathrm{ssim}}\,\mathcal{L}_{\mathrm{SSIM}}
  + \lambda_{\mathrm{bg}}\,\mathcal{L}_{\mathrm{bg}},
  \label{eq:total_loss}
\end{equation}
where $\mathcal{L}_1$ is the per-pixel photometric loss, and $\mathcal{L}_{\mathrm{SSIM}}$ is the D-SSIM structural similarity loss between the rendered and ground-truth images. We leave the SSIM loss scaling factor $\lambda_{\mathrm{ssim}} = 0.2$ unchanged at its default value from~\cite{kerbl20233dgs}. Lastly, following~\cite{kerbl20233dgs}, every eighth image is held out for evaluation during training.

%% file: sections/04_examples.tex
\section{Experiments}
\label{sec:experiments}

\begin{figure}[t]
\centering
\includegraphics[width=\columnwidth]{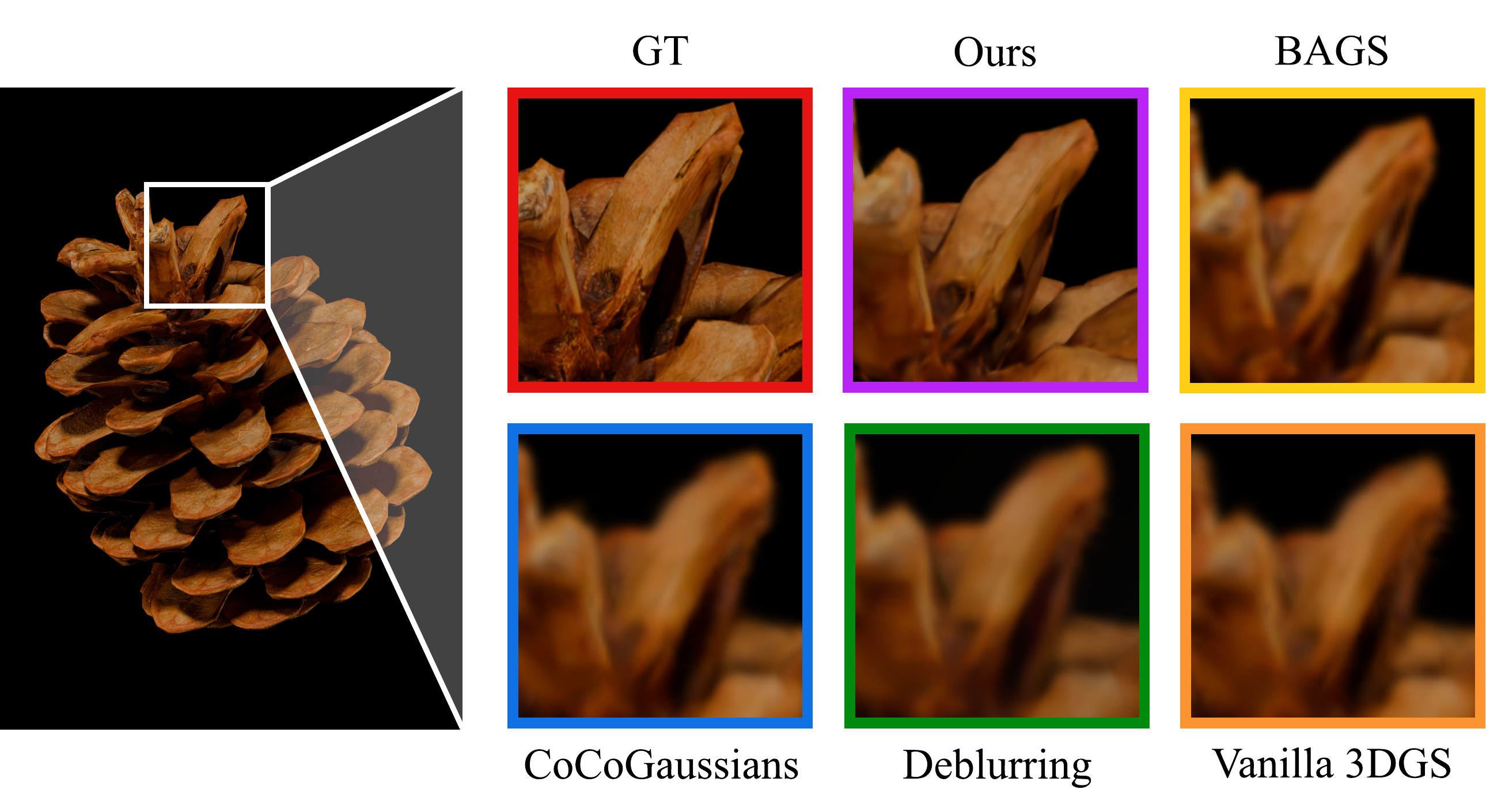}
\vspace*{-0.5cm}
\caption{Qualitative comparison on a pinecone from our synthetic dataset.}
\label{fig:comparison}
\vspace*{-0.3cm}
\end{figure}



\begin{figure*}[t]
  \centering
  \includegraphics[width=\textwidth]{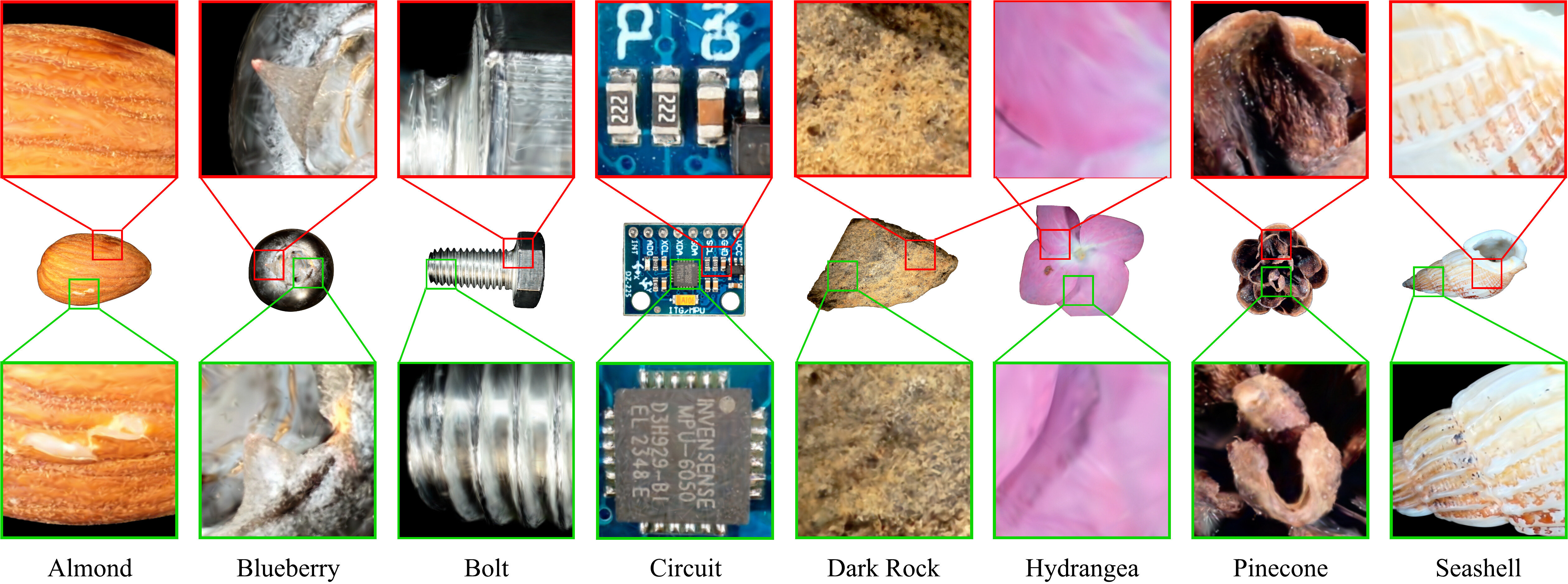}
  \vspace*{-0.5cm}
  \caption{Qualitative real-world reconstruction results.}
  \label{fig:real_world_results}
  \vspace*{-0.3cm}
\end{figure*}
We now experimentally validate the MicroSplat pipeline. First, we ablate the object-centering step to see how well it can localize and keep an object centered during data collection. Second, we test various SfM initialization methods to see which version produces the best results.

We also benchmark against vanilla 3DGS \cite{kerbl20233dgs} (trained on images that neither focus-stacked or masked) and three state-of-the-art deblurring approaches in 3DGS: BAGS \cite{peng2024bags}, CoCoGaussian \cite{lee2025cocogaussian}, and Deblurring 3DGS \cite{lee2024deblurring3dgs}.

\subsection{Runtime}
On average, the full MicroSplat pipeline takes 134 minutes per object on a single NVIDIA L40S: 2 minutes for object-centering, 39 minutes for data collection, 33 minutes for pre-processing (focus-stacking and masking), and 60 minutes for the reconstruction (COLMAP and 3DGS).


\subsection{3DGS Training Parameters}
We follow the original 3DGS~\cite{kerbl20233dgs} training configuration,
optimizing for 30{,}000 iterations from the COLMAP sparse point cloud with
input images downsampled by a factor of 4 from the microscope's
$1920\times1080$ resolution, and use identical settings for both the real and
synthetic datasets.

We deviate from the defaults in two places, both chosen to help with suppressing floater
artifacts: the opacity pruning threshold $\tau_\alpha$ is doubled from $0.05$
to $0.10$, and the 2D
gradient growth threshold $\tau_\nabla$ also doubled from $2\times10^{-4}$ to
$4\times10^{-4}$.

\subsection{Datasets}

\subsubsection{Real-World Dataset}
To qualitatively evaluate the system, we selected eight samples (almond, circuit, blueberry, bolt, rock, pinecone, hydrangea, and seashell) spanning surface appearance (matte to slightly specular), brightness, geometric complexity, texture frequency, and depth extent.
Each sample's maximum width is 15-\SI{27}{\milli\meter}.

\subsubsection{Synthetic Dataset}
We quantitatively evaluate focus stacking in simulation because no reliable ground truth exists for the real captures: every real image is degraded by the shallow DoF the pipeline is designed to correct, so there is no in-focus reference to score against.

We generate a synthetic Blender dataset of 8 samples, modeled after the same objects as the real-world set. Camera poses are inherited from the real-world almond COLMAP reconstruction, which recovers a single refined pose per viewpoint — that of the focus-stacked image. Since each stack sweeps 40 mm over 30 images, we recover the remaining 29 poses by applying the corresponding axial offset from the selected stack index. Each synthetic model is then scaled and positioned to sit within these reconstructed scan points.


Since the microscope's manufacturer does not specify its aperture or focus distance, we recover it from the 42 images already captured for intrinsic calibration (\Cref{sec:method}), fitting a thin-lens defocus model to per-corner blur and depth (from PnP) on the well-conditioned views, yielding a median working f-number of $N=6.2$ and focus distance of \SI{104}{\milli\meter}.

\begin{table}[H]
\centering
\footnotesize
\setlength{\tabcolsep}{3pt}
\begin{tabular*}{\columnwidth}{@{\extracolsep{\fill}} cc rrrrrr @{}}
\toprule
 & & \multicolumn{2}{c}{PSNR $\uparrow$} & \multicolumn{2}{c}{SSIM $\uparrow$} & \multicolumn{2}{c}{LPIPS $\downarrow$} \\
\cmidrule(lr){3-4} \cmidrule(lr){5-6} \cmidrule(lr){7-8}
Poses & Intrinsics & Mean & Std. & Mean & Std. & Mean & Std. \\
\midrule
No init. & No init. & 19.42 & 5.04 & 0.672 & 0.145 & 0.389 & 0.186 \\
Init.    & No init. & \cellcolor{rank2} 21.65 & 2.19 & \cellcolor{rank2} 0.701 & 0.076 & \cellcolor{rank3} 0.365 & 0.128 \\
No init. & Init.    & \cellcolor{rank3} 20.81 & 4.42 & \cellcolor{rank3} 0.697 & 0.157 & \cellcolor{rank2} 0.363 & 0.201 \\
Fixed    & No init. & 16.39 & 2.19 & 0.593 & 0.104 & 0.554 & 0.130 \\
Fixed    & Fixed    & 17.35 & 1.53 & 0.592 & 0.092 & 0.507 & 0.116 \\
Fixed    & Init.    & 16.58 & 2.15 & 0.607 & 0.093 & 0.518 & 0.128 \\
\midrule
Init.    & Init.    & \cellcolor{rank1} 26.37 & 0.28 & \cellcolor{rank1} 0.911 & 0.002 & \cellcolor{rank1} 0.123 & 0.003 \\
\bottomrule
\end{tabular*}
\vskip\abovecaptionskip\relax
\caption{SfM initialization ablation (averaged across real objects). For robot poses and calibrated intrinsics, No init.\ denotes bundle adjustment by COLMAP without a prior, Init.\ denotes initialization from the prior followed by refinement, and Fixed denotes freezing to the prior. The bottom row is our configuration. Top three per metric are colored red, orange, and yellow, respectively.}
\label{tab:quantitative_metrics}
\end{table}

\subsection{Ablation Study}

\subsubsection{Object-Centering Experiment}
We evaluate the object-centering algorithm on a spherical mocap marker, whose center is well-defined from any viewing angle. After running the centering and data-collection stages, we report the pixel distance between the sample's centroid and the true image center across all scan poses. For each viewpoint, we focus-stack the images as in the main pipeline, then manually draw a bounding box on the focus-stacked image and take its center.

The results show an average pixel error of \SI{98}{\pixel}, which is consistent with the combined effect of the pipeline's known error sources.
The \SI{1.07}{\milli\meter} hand-eye calibration residual and the \SI{1.0}{\milli\meter} manual jogging increment suggest the centering accuracy is limited by the mount's imperfect rigidity and discrete jogging precision rather than the ray-intersection procedure itself.

\subsubsection{SfM Initialization Priors}

To assess the robot's reported poses and our calibrated intrinsics as SfM priors, we compare configurations in which each prior is unused, used to initialize bundle adjustment, or fixed, evaluated on the real-world dataset (\Cref{tab:quantitative_metrics}).

Fixing the robot poses performs worst (PSNR $<$\SI{18}{\decibel}) regardless of how intrinsics are handled. Although the arm itself is precise, the mount's imperfect rigidity leave the derived camera poses too imprecise for high-fidelity rendering without refinement. Initializing with either prior alone improves over using neither ($+$\SI{2.23}{\decibel} for poses, $+$\SI{1.39}{\decibel} for intrinsics), but initializing with both yields the largest gain ($+$\SI{6.95}{\decibel}) and reduces the standard deviation across objects from \SI{5.04}{\decibel} to \SI{0.28}{\decibel}, indicating the priors are complementary.

\subsection{Comparisons}
\label{sec:comparisons}



To compare against baseline methods, our method is initialized with the robot's reported poses and calibrated intrinsics and lets COLMAP refine them, mirroring how the pipeline runs on real data. Each baseline uses the same parameters from its publication, with minor adjustments to accommodate our sample sizes. For these methods, SfM on the defocused inputs produced obvious pose errors, so we instead supply known intrinsics and frozen ground-truth poses. This is an advantage our method does not receive.
We run the blur-aware baselines without masks, as published, since they model depth-dependent defocus; to give them realistic defocus to model, our synthetic scenes include a background modeled after our real-world workspace.

Our method consistently outperforms the blur-aware methods (\Cref{tab:method-comparison}, \Cref{fig:comparison}), as they must infer sharp detail, while the arm's controlled focal sweep ensures it is measured directly.
The gap is especially large because many of our frames are fully out of focus.
Deblurring 3DGS and CoCoGaussian realize blur through the Gaussians themselves, so any blur they cannot represent is absorbed into geometry, leaving them below even vanilla 3DGS.
BAGS applies its blur as a 2D convolution on the final render, leaving geometry untouched, and performs best among the baselines.

\subsection{Real-World Results}
The synthetic dataset controls lighting, exposure, and white balance; the real captures have none of these guarantees, so these results demonstrate the pipeline's ability to extrapolate to our low-cost microscope. As shown in \Cref{fig:real_world_results}, many of the reconstructions are difficult to distinguish from photographs. Intricate detail survives: the peeling skin on the almond, the flap on the blueberry, and the even the text on the circuit board, whose characters are roughly 0.4 mm wide. The hydrangea is the weakest result. Its petals are thin, which the volumetric 3D Gaussian representation is known to handle poorly~\cite{huang2024twodgs}, and its uniform coloring yields sparse SfM initialization.

%% file: sections/05_conclusion.tex
\section{Conclusion}

We presented MicroSplat, a robotic pipeline that turns a low-cost consumer microscope on a KUKA iiwa14 R820 into an autonomous scanner for centimeter-scale objects. After calibration and a brief object-centering step, it runs unattended from capture to a finished 3DGS model with sub-millimeter detail. We also release the real and synthetic datasets used in this evaluation.

The system is still limited in many ways; it is not fully autonomous, and our reconstructions do not provide full \SI{360}{\degree} coverage.
Looking further ahead, we believe the solutions we developed for lighting inconsistency, pose inaccuracy, and non-uniform backgrounds will be useful for a future robotic system capable of scanning and reconstructing objects in the wild without disturbing them.


